\documentclass[10pt,twocolumn,letterpaper]{article}

\usepackage{cvpr}
\usepackage{times}
\usepackage{epsfig}
\usepackage{graphicx}
\usepackage{amsmath}
\usepackage{amssymb}
\usepackage{paralist}
\usepackage{xfrac}
\usepackage{verbatim}

\usepackage[captionskip=0.5mm, farskip=0.5mm]{subfig}

\usepackage{multirow}

\newcommand{\thdr}[1]{\textbf{\small{#1}}}
\newcommand{\tcnt}[1]{{\small{#1}}}

\graphicspath{{./gfx/}}

\usepackage[pagebackref=true,breaklinks=true,letterpaper=true,colorlinks,bookmarks=false]{hyperref}

\cvprfinalcopy 

\def\cvprPaperID{7} 

\ifcvprfinal\fi
\begin{document}

\title{Parametric annealing: a stochastic search method for human pose tracking}

\author{Prabhu Kaliamoorthi\\
Nanyang Technological University\\
Singapore\\
{\tt\small prab0009@e.ntu.edu.sg}
\and
Ramakrishna Kakarala\\
Nanyang Technological University\\
Singapore\\
{\tt\small ramakrishna@ntu.edu.sg}
}

\maketitle

\begin{abstract}
Model based methods to marker-free motion capture have a very high computational overhead 
that make them unattractive. In this paper we describe a method that improves on existing
global optimization techniques to tracking articulated objects. Our method 
improves on the state-of-the-art  Annealed Particle Filter (APF) by reusing 
samples across annealing layers and by using an adaptive parametric density for diffusion. 
We compare the proposed method with APF on a scalable problem and study how the two 
methods scale with the dimensionality, multi-modality and the range of search.
Then we perform sensitivity analysis on the parameters of our algorithm and show 
that it tolerates a wide range of parameter settings. We also show results on tracking human 
pose from the widely-used Human Eva I dataset. Our results show that the proposed 
method reduces the tracking error despite using less than 50\% of the computational 
resources as APF. The tracked output also shows a significant qualitative improvement 
over APF as demonstrated through image and video results. 
\end{abstract}

\section{Introduction}

Tracking an articulated object like a human body with many degrees of freedom is an active research area in the vision community. A large body of research on human pose tracking and estimation suitable to different application areas is found in the literature \cite{Survey2}.  The model-based generative method to tracking humans  \cite{GenerativeBayesianFormulation,Deutscher2000,Deutscher2001,Sminchisescu03estimatingarticulated,cham:a,RaskinRR07a,LearnMotionCorr,Gall_Rosenhahn_Brox_Seidel_2008} is a critical method with applications in animation, sports and medical motion analysis. Relying on a kinematic model to support the tracked hypothesis it remains one of the most accurate ways to track a human. However the accuracy comes with a significant computational overhead.

Model based generative methods model a human by a kinematic tree controlled by a fixed set of parameters. The imaging process is modeled by a projection operation and pose estimation is formulated as a state estimation problem that aims to minimize the disparity between the actual observation and the generated image as a function of the parameters of the kinematic model. Every evaluation of the disparity function incurs a significant computational overhead which can be considered to be of the order of the resolution and the number of cameras used for tracking. 

Existing literature on the method either approach the problem as that of a Bayesian filtering problem \cite{Chen_2003} or as an optimization problem. Since the observation is not a random vector correlated with the hidden state, as is the case in filtering problems, Bayesian methods either resort to local optimization \cite{cham:a,Sminchisescu03estimatingarticulated} to recover and track the modes or choose a nonparametric method like particle filter \cite{GenerativeBayesianFormulation}. In practice, however, both these methods need a significant number of evaluations of the disparity function, resulting in a high computational overhead to track. 

Alternate approaches \cite{Gall_Rosenhahn_Brox_Seidel_2008,RaskinRR07a,ApfIjcv} formulate the problem as a search for the global mode of the likelihood assuming the disparity to be the negative log likelihood. This line of techniques originate from the Annealed Particle Filter (APF) \cite{ApfIjcv}. APF is a generic procedure applicable to several types of input data including silhouettes from video streams and 3D reconstructions of human. Moreover it does not rely on learned prior dynamics \cite{HumanEvaII}, hence it remains an attractive algorithm and is largely considered to be the state-of-the-art in model-based human pose tracking.

APF reveals that annealing can be a powerful tool when applied to high dimensional multi-modal problems. However, we observe that the full potential of annealing is not realized by the APF. Since the cost of each likelihood evaluation is very high, there is a need to extract as much information as possible from each sample. Our procedure, which we refer to as parametric annealing (PA), is aimed at reducing the total number of samples by reusing samples across annealing layers and by using an adaptive parametric density to generate new samples. Our experiments show that it is capable of tracking more accurately with less than 50\% of the number of samples required by APF. In this paper, we study various properties of our method including how it scales with the number of modes and dimensions, and also examine its sensitivity to parameter settings. A preliminary version of our paper appears in \cite{hau3d2012}.

The rest of the paper is organized as follows. Section 2 provides a background introduction to annealed particle filters. Section 3 highlights the problems with APF and motivates the need for our method. Section 4 provides a detailed description of our method. Section 5 compares the proposed method with APF on a simple scalable problem and performs sensitivity analysis on the parameters of our method. Following that, we provide a summary of the tracking results using data from Human Eva I dataset. Section 6 indicates the future work and concludes the paper.

\begin{figure}
	\begin{center}
		{\includegraphics[width=8cm, height=2.25cm, clip=true, trim=3.5cm 1.35cm 5.8cm 24.65cm]{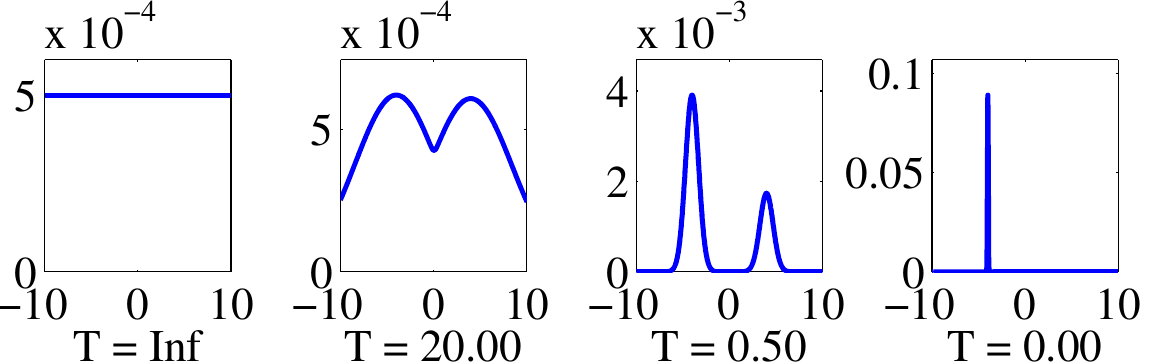}}
	\end{center}
	\caption[Effect of annealing]{The plots shows the effect of annealing on a Mixture of Gaussians as the temperature T is varied from $\infty$ to 0. Annealing temperature is displayed below the plot.}
	\label{fig:anneal}
\end{figure}

\section{Background}
\label{sec:apfintro}
In this section we review the background of annealed particle filters to motivate the need for our method and to enable the exposition of it.

Particle filter \cite{condensation} is a well-known technique in computer vision for tracking. It propagates the uncertainty about the tracked object non parametrically using a set of weighted samples. Let $x_t$ and $z_t$ be random variables corresponding to the state of the object and observation at time $t$. Let $\psi$ be the set of $N$ samples $x_i$ and their corresponding normalized weights $\pi_i$ that represent the distribution of $x_t$. The notation $z_{t:1}$ is used to indicate the set of observations till time $t$. Formally
\begin{equation}
	p(x_t | z_{t:1}; \psi) = \sum_{i=1}^N \pi_i\,\delta(x_t - x_i)
\end{equation}
Applying the first order Markov and the sensor assumptions \cite{condensation} one can obtain the posterior at time $t+1$ by multiplying the likelihood with the prior, where the prior is obtained by convolving posterior from time $t$ with the motion prior. Assuming the motion prior $p(x_{t+1}|x_t)$ to be a combination of a function $f(x_{t})$ and Gaussian uncertainty of covariance $\Sigma$, results in a prior density shown below
\begin{equation}
\label{eq:prior}
	p(x_{t+1} | z_{t:1}; \psi) = \sum_{i=1}^N \pi_i\,\mathcal{N}(f(x_i); \Sigma)
\end{equation}
Particle filters use the prior density as a proposal density for importance sampling\cite{MacKay} from the likelihood. Multiplying the likelihood by the prior provides the posterior at time $t+1$. This results in a four step procedure for tracking i.e. re-sample, drift, diffuse and evaluate\cite{condensation}. The first three steps can be shown to generate samples from the prior density (\ref{eq:prior}), and the last step can be shown to perform importance sampling and the necessary multiplication to obtain the posterior. The algorithm hinges on the critical assumption that the prior density has a good overlap with the likelihood for the importance sampling to be effective.

Annealing has the effect of transforming any distribution from a uniform distribution to a delta function in the global maximum as the temperature is changed from $\infty$ to 0. Figure \ref{fig:anneal} shows the effect of annealing on a multimodal distribution. Formally, the process of annealing can be expressed as below
\begin{equation}
	f(x, T) \propto {p(x)}^{\frac{1}{T}}
\end{equation}
where $p(x): \mathbb{R}^d \mapsto \mathbb{R}$, is a non-negative function with a finite integral and T is the annealing temperature. 

APF\cite{ApfIjcv} incorporates annealing into particle filtering by exploiting the fact that a distribution at a higher temperature effectively has a wider support (as can be observed in Figure \ref{fig:anneal}), and hence could be used as a proposal distribution for importance sampling from the same distribution at a lower temperature. Hence in a single iteration (that corresponds to a time interval $t$), it performs several particle filtering steps referred to as layers. It assumes $f(x_t)$ to be $x_t$ in successive layers (since the distribution is not translated by annealing) and the variance to be gradually decreasing by a fraction $\alpha^m$ (where $m$ is the annealing layer and $\alpha < 1$) to account for the narrow and peaked objective that results from annealing (see Figure \ref{fig:anneal}). Annealing is performed by a simple schedule by controlling the number of unique resampled particles to be approximately half of that which exist in the previous layer. Consequently, at the end of a fixed number of layers, the set of samples that represented a multimodal density in the start represent a delta distribution in the end. Hence the expected state of the set of samples approximates the global maximum with high accuracy. The high level steps done in APF are as follows.
\begin{enumerate}
\item $N$ samples $x_i, i \in \{1,\dots,N\}$ are obtained by sampling from $\mathcal{N}(\hat{x}_{t}; \Sigma)$, where $\hat{x}_{t}$ is the estimate of the state in last iteration.
\item The likelihood $y_i$ is evaluated at sample $x_i$. 
\item A suitable annealing temperature $T_m$ that ensures only 50\% samples will survive resampling is estimated. 
\item Normalized sample weights are obtained. $\pi_i = \sfrac{y_i^{\beta_m}}{\sum_{i=1}^{N} y_i^{\beta_m}}$, where $\beta_m = \sfrac{1}{T_m}$
\item N new samples are generated by sampling (resample and diffuse) from $\sum_{i=1}^N \pi_i\,\mathcal{N}(x_i; \alpha^m\Sigma)$. 
\item Repeat from step 2 for a fixed number of layers.
\item The global maximum is estimated as the expected state of the weighted samples.
\end{enumerate}

The critical aspect of APF is that it is capable of tracking with significantly fewer number of samples. Studies done on human pose tracking show that APF is capable of tracking with 1000 \cite{HumanEvaII} samples in comparison to 10,000 samples that are required by particle filters \cite{GenerativeBayesianFormulation}.

\section{Motivation}
\label{sec:motivation}
The APF algorithm generates N new samples in each layer. However, samples from previous layers are discarded. Since the evaluation of likelihood incurs a significant overhead, one would want to retain the relevant samples from previous layers. The most natural way to do this would be to simply retain the samples $x_i$ along with the corresponding likelihoods $y_i$, and raise them to a new annealing temperature $T_m$ for the layer before normalization. However, this does not work well in practice.
\begin{figure}
	\centering
	\subfloat[Discard]{\includegraphics[width=2.5cm, height = 1.6cm, clip=true, trim=4.6cm 15.85cm 11cm 9.4cm]{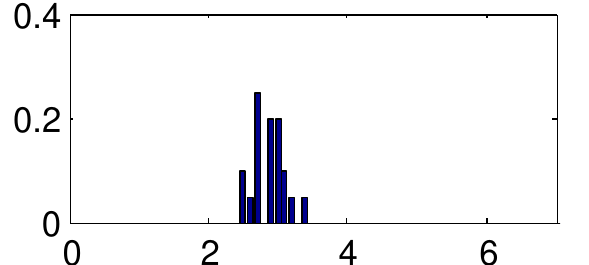}\label{fig:apftest1_a}}
	\hspace{1mm}
	\subfloat[Retain]{\includegraphics[width=2.5cm, height = 1.6cm, clip=true, trim=4.6cm 12.5cm 11cm 12.75cm]{apftest.pdf}\label{fig:apftest1_b}}
	\hspace{1mm}
	\subfloat[Improved]{\includegraphics[width=2.5cm, height = 1.6cm, clip=true, trim=4.6cm 9.1cm 11cm 16.1cm]{apftest.pdf}\label{fig:apftest1_c}}
	\\
	\caption[Test demonstrating effects of reusing samples]{Test demonstrating effects of reusing samples. The subfigures a, b and c correspond to the configuration that discards samples, the one that retains them and the one that retains along with a deep annealing schedule. It can be observed that reusing makes the search less effective (b), which is overcome by augmenting it with a deep annealing schedule (c).}
	\label{fig:apftest1}
\end{figure}

To demonstrate the effect of simply reusing, we conducted an experiment. We considered the log likelihood to be a quadratic, $-\frac{1}{2}x^Tx$, where $x \in \mathbb{R}^{30}$. To verify how well APF searches through the state space we started from a fixed state and searched for the optimum. We compared two configurations of APF: the original configuration that discards samples, and a second configuration which retains them by raising the likelihood to the appropriate $\beta_m$ for the layer before normalization. To measure how effectively the methods locate the optimum, we obtained the distribution of the L2 norm of the state estimated by APF. The histograms of the L2 norm are shown in Figure \ref{fig:apftest1_a} \& \ref{fig:apftest1_b}. It can be observed that when samples are discarded, the state estimate is closer to the true optimum. 

\begin{figure}
	\centering
	\subfloat[Original]{\includegraphics[width=4cm, height=1.6cm, clip=true, trim=4.5cm 15.7cm 3.8cm 10.8cm]{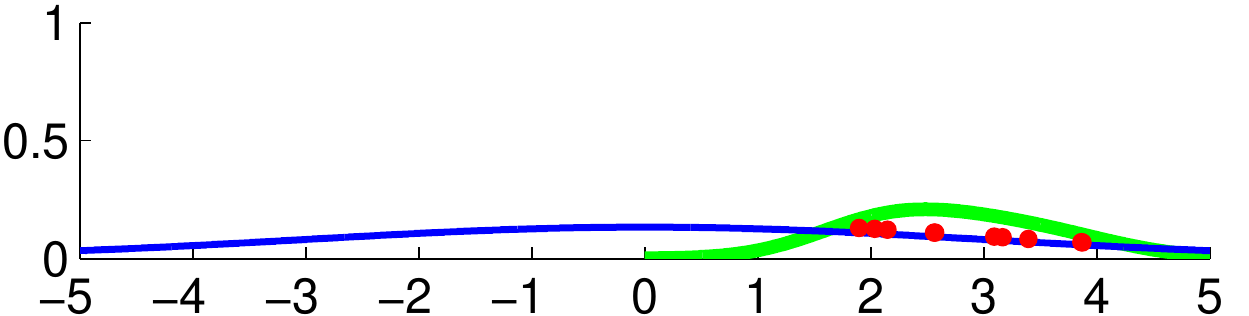}\label{fig:annealexp_a}}
	\subfloat[Annealed T=10]{\includegraphics[width=4cm, height=1.6cm, clip=true, trim=4.5cm 12.4cm 3.8cm 14.1cm]{annealexp.pdf}\label{fig:annealexp_b}}
	\\
	\caption[Toy example demonstrating the effect of annealing schedule]{Example demonstrating the effect of annealing(color). It can be seen how annealing shifts the probability mass towards regions that improve the objective.}
	\label{fig:annealexp}
\end{figure}

The test demonstrates the problems with retaining samples. The critical insight, however, is that the problems with reusing samples are more than overcome by choosing a ``deep'' annealing schedule that consistently reduces the annealing temperature, rather than a simple schedule which ensures 50\% samples survive resampling. The precise formulation of a deep annealing schedule as a power law, which is a novel aspect of our method, is given in Section \ref{sec:parannealoverview}. The improvement is shown in the histogram in Figure \ref{fig:apftest1_c} which augments retaining samples with our proposed deep annealing schedule. The result can be best explained by analyzing another simple example. Figure \ref{fig:annealexp_a} shows a 1D Gaussian density in blue. A set of weighted samples were obtained by diffusing samples around $x=3$. The samples are shown in red and the kernel density approximated by the samples using a Gaussian kernel is shown in green. If the samples were to be re-sampled and diffused with the same Gaussian kernel, it is equivalent to generating samples from the density shown in green. Figure \ref{fig:annealexp_b} shows the effect of annealing on the kernel density. It can be observed that annealing effectively shifts the kernel density mass towards regions that improve the search objective. Consequently, when enabled with a suitable annealing schedule, new samples are generated in areas of the state-space that improve the objective. Expressed differently, a good annealing schedule ensures that the worthless samples which end up in areas of the state-space with poor objective don't adversely impact the search process by quickly driving their normalized weights to zero in successive layers.

In addition, one would expect to have a slow annealing schedule when tracking articulated objects since their likelihoods are multimodal and fast annealing is known to converge to local optima. APF can be interpreted as generating new samples from a kernel density with a Gaussian kernel of gradually reducing noise covariance. It is well-known that the kernel density approximated by a set of samples is highly sensitive to the type of kernel. Therefore such a fixed diffusion schedule is very sensitive to the parameters, more so when the annealing is slow and longer. This is evident from the two-fold reduction \cite{Deutscher2001} in the number of samples needed to track when using an adaptive kernel. We take the idea of adaptive diffusion density a step further, and use an intermediate parametric form which is inferred from the set of samples to act as a substitute for the kernel density i.e. instead of resample and diffuse with a kernel, new samples are generated from a parametric model. The motivation is that, with an intermediate parametric form, the density would adapt even better. 

\begin{figure}
	\begin{center}
	\centering
	\subfloat{{\includegraphics[width=4cm, height=2.25cm, clip=true, trim=80pt 200pt 110pt 305pt]{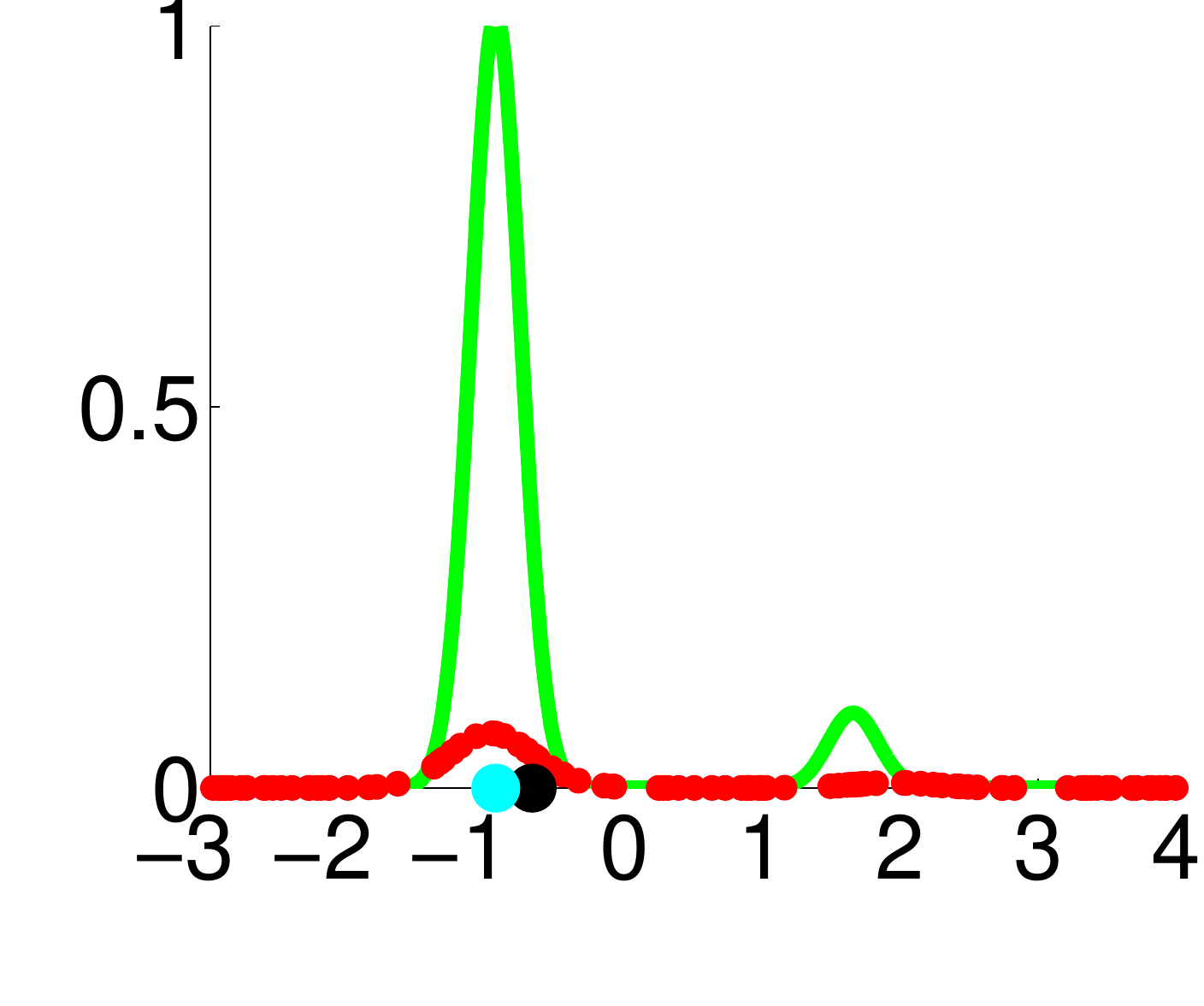}}}
	\caption[Toy example demonstrating the effect on estimate]{Example demonstrating the effect of a parametric form on the state estimate. The samples are shown in red, and an inferred parametric form is shown in green. The expected state of all samples, and the maximum of the parametric form, are shown in black and cyan respectively. It can be seen how the expected state is a worse estimate of the true optimum at -1. }
	\label{fig:estimatedemo}
	\end{center}
\end{figure}

Inferring an intermediate parametric form has an added advantage when the set of samples in the last layer of annealing still represent a multimodal density. Such a scenario can be observed in Figure \ref{fig:estimatedemo}, it shows a set of samples in the last layer of annealing in red. A parametric density inferred from the set of samples is shown in green. The parametric density shows that the samples represent a multimodal density. It can be observed that the expected state of such a sample set deviates from the global maximum, and a better estimate is obtained from the parametric form. These considerations motivate the need for an improved procedure, which we refer to as parametric annealing. We describe our method in detail in the next section.

\section{Parametric Annealing}
\label{sec:parannealoverview}
Let $P(x)$ $(x \in \mathbb{R}^d)$ be a general multi-modal distribution, and let $\psi$ be the set of $N$ weighted samples $(x_i, \pi_i)$ that represent $P(x)$. Formally,
\begin{equation}
	P(x) \approx p(x; \psi) = \sum_{i=1}^N \pi_i\,\delta(x - x_i)
\end{equation}
Let $q(x; \theta)$ be a Mixture of Gaussians (MoG), parametrized by $\theta$ that approximates $p(x; \psi)$. We estimate the parameters of $q(x; \theta)$ using Expectation Maximization (EM) adapted to include the sample weights $\pi_i$. Formally,
\begin{equation}
	q(x; \theta) \approx p(x; \psi) \approx P(x)
\end{equation}

With this terminology, the high level steps done in each iteration of our algorithm are now described. Subsequently, the steps are compared to the APF, and the differences are highlighted.

\begin{enumerate}
\item $N^0$ samples $x_i, i \in \{1,\dots,N^0\}$ are obtained by sampling from $\mathcal{N}(\hat{x}_{t}; \Sigma)$, where $\hat{x}_{t}$ is the estimate of the state in last iteration.
\item The likelihood $y_i$ is evaluated for new samples $x_i$.
\item A suitable annealing temperature $T_m$ is estimated using a deep annealing schedule.
\item Normalized sample weights $\pi^m_i = \sfrac{y_i^{\beta_m}}{\sum_{i=1}^{N^m} y_i^{\beta_m}}$, where $\beta_m = \sfrac{1}{T_m}$ are estimated.
\item A parametric MoG approximation $q(x; \theta^m)$ is inferred from the weighted samples $\psi$ using one EM iteration.
\item $C$ new samples $x_i$ are generated from the parametric model and combined with old samples.
\item Repeat from step 2 for $M$ layers to simulate a very slow annealing.
\item The global maximum is estimated from the parametric model.
\end{enumerate}
The main novelty of our procedure over APF is twofold.
\begin{itemize} 
\item We don't discard samples in step 6, and moreover, we use a deep annealing schedule to enable effective search in step 3. 
\item We use a parametric model in steps 5 \& 6 instead of the kernel diffusion density, and obtain the estimate from the parametric form in step 8. 
\end{itemize}
Below, we provide details of the method, which may be skipped by those readers who want to consider the results in the next section.  

The high level steps above include some of the parameters of the algorithm. The superscript $m$ in any parameter indicates that it is a function of the layer, $m \in (1, \dots, M)$, where $M$ is the number of annealing layers. The parameter $N^m$ represents the total number of samples in layer $m$. As opposed to APF where the number of samples in each layer remain fixed, the number of samples in our method grows in successive layers since we don't discard samples. As we introduce $C$ new samples in a layer and we start with $N^0$ samples, the number of samples in each layer $N^m$ is $N^0 + mC$, bringing the total number of samples for an iteration to $N^0 + MC$.

We fix the number of mixture components in the MoG $q(x; \theta)$ to $C$, which is also the number of new samples introduced in a layer. Consequently, the parameters $\theta^m$ is the set of $C$ means $(\mu^m_c)$, covariances $(\Sigma^m_c)$ and the mixture weights $(\phi^m_c)$. The subscript $c$ and the superscript $m$ indicates that these parameters are dependent on the specific mixture component $c$ and the layer $m$. The ``E \& M'' steps \cite{Bishop:2006:PRM:1162264} in the EM algorithm are well known.  A regularizer $\sfrac{\xi_m}{\phi^m_c}$ in included in the M step update for $\Sigma^m_c$ to ensure that the covariance matrices are full rank. Since we are also annealing the samples while inferring the parametric form we find that a gradually reducing regularizer is more effective than a fixed regularizer.

The annealing schedule is a critical part of our algorithm. We use a measure similar to particle survival rate $\alpha$ defined in \cite{ApfIjcv} to control the annealing schedule. The parameter $T_m$ which is the annealing temperature for the layer $m$ can be estimated from a predefined sequence of $\alpha_m$ by any local optimization technique. The parameters $\alpha_m$ and $\xi_m$ (regularizer) define the annealing schedule. We define these parameters by the following power law
\begin{equation}
	\xi_m = \alpha_m \xi \xi' \alpha_m, \quad \alpha_m = \eta \lambda^m
\end{equation}
Where $\xi$ is a $d\times d$ diagonal matrix with each diagonal element equaling half the maximum change in state estimate along that dimension. The significant difference between our schedule and that used by APF is that APF sets the parameter $\alpha_m$ to 0.5 for all $m$. Thus $\eta \in (0, 1)$, $\lambda \in (0, 1)$, $N^0$, $M$ and $C$ are the external parameters used by our method. These parameters are dependent upon the state space dimension $d$ and the density $P(x)$. 

\section{Analysis and comparison}
\subsection{Scaling properties}
In this section we analyze how our method scales with the dimensionality, multi-modality and the range of search by comparing it with APF on a simple scalable problem. The problem used for the comparison is to let the two stochastic procedure search for the optimum state in a MoG. Inspired by \cite{Blei05variationalinference}, we used a generative method to model the MoG used as the search objective. We assume the search objective $\theta$ (which is a MoG) to be a random draw from a distribution $P(\Theta)$. The distribution $P(\Theta)$ is sampled using a generative method i.e., the mixture weights are drawn from a Dirichlet distribution of uniform prior, the inverse covariances are drawn from a Wishart distribution with $d$ dimensional identity as scale matrix and the mean vectors are drawn from a $d$ dimensional Gaussian with a scaled identity matrix (by scale $s$) as covariance and zero mean. 

\begin{figure}
	\centering
	\subfloat{\includegraphics[width=2.8cm, height=2.3cm, clip=true, trim=4.9cm 8.6cm 11cm 8.9cm]{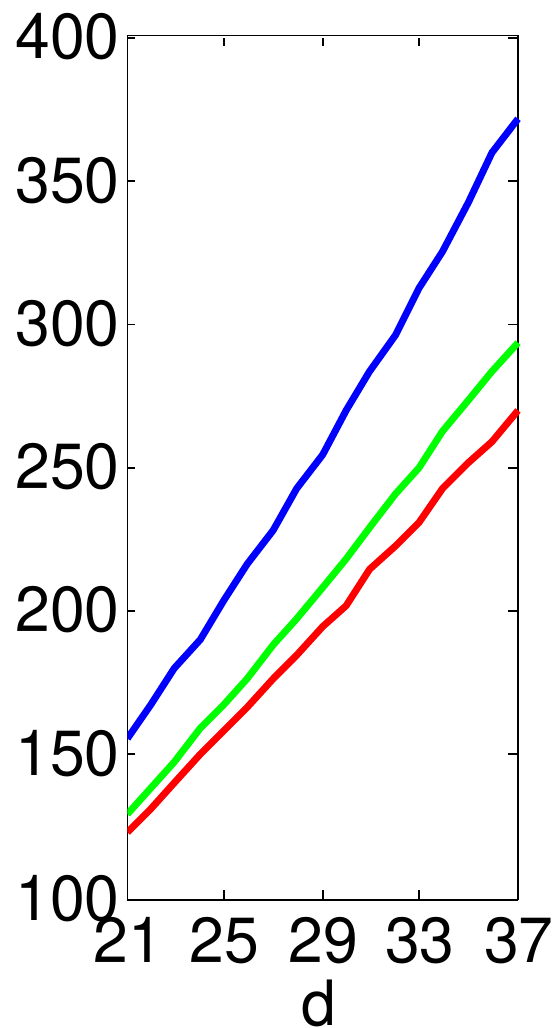}}
	\subfloat{\includegraphics[width=2.8cm, height=2.3cm, clip=true, trim=4.9cm 8.6cm 11cm 8.9cm]{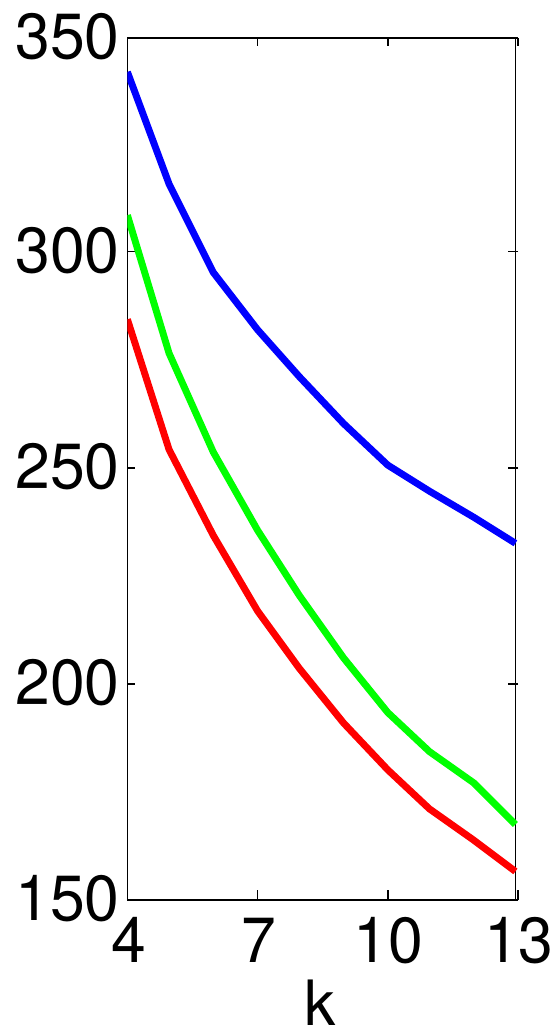}}
	\subfloat{\includegraphics[width=2.8cm, height=2.3cm, clip=true, trim=4.7cm 8.6cm 11cm 8.9cm]{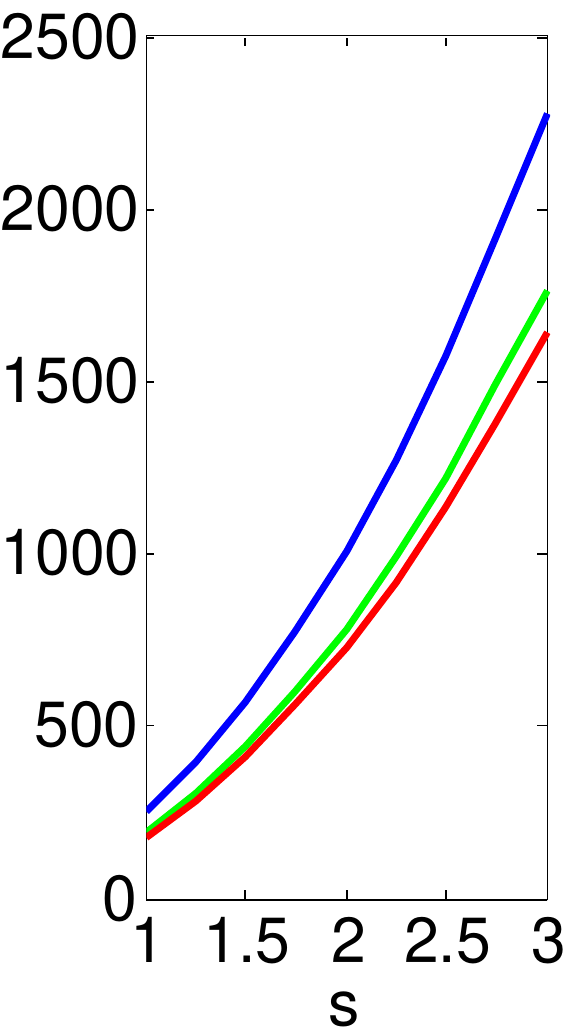}}
	\\
	\subfloat{\includegraphics[width=2.8cm, height=2.3cm, clip=true, trim=11.42cm 8.6cm 4.48cm 8.9cm]{scaledim.pdf}}
	\subfloat{\includegraphics[width=2.8cm, height=2.3cm, clip=true, trim=11.42cm 8.6cm 4.48cm 8.9cm]{scalecomp.pdf}}
	\subfloat{\includegraphics[width=2.8cm, height=2.3cm, clip=true, trim=11.12cm 8.6cm 4.48cm 8.9cm]{scalerange.pdf}}
	\\
	\caption[Scaling properties of pa]{Scaling properties of the PA (blue) in comparison to the APF 1000 (green) and APF 500 (red). The plots shows the mean (top row) and deviation (bottom row) of improvement in log likelihood for different values of dimension $d$, number of components $k$ and range parameter $s$ respectively. It can be observed that our method improves the likelihood more than APF, and that the deviation in improvement is almost the same.}
	\label{fig:scaleprop}
\end{figure}

The variable $d$, the number of mixture components $k$ and the scale of the covariance $s$ control the dimensionality, multimodality and search range of the objective. We made both the procedures start from the origin of the euclidean space $\mathbb{R}^d$ and search for the optimum. We considered the difference in the log likelihood between the end state and the start state to be a random variable $I$ (for improvement in the log likelihood) that acts as an indicator of how well the procedure performs. Since the procedures are stochastic in nature we ran the procedure several times on different samples from $P(\Theta)$ and considered the mean and standard deviation of $I$ to be indicative of the performance.

Figure \ref{fig:scaleprop} plots the scaling properties of the proposed method in comparison to the APF. We compared our method to two configurations of APF. The configurations are APF 1000 ($M=5, N=200$) and APF 500 ($M=5, N=100$), where $M$ and $N$ are annealing layers and particles per layer respectively. Our method was configured with 438 samples ($438=N^0 + M*C = 150+24*12$). The mean and standard deviation of the random variable $I$ is plotted against various values of dimension $d$, components $k$ and scale factor $s$. Higher mean value of $I$ and lower deviation are better. The result might seem to show that all procedures somehow perform better in higher dimensions and as the range increases, but this is not true. It is caused by the vast change in the log likelihood in higher dimensional Gaussian mixtures and over long search ranges. However, it can be observed that, in all three tests, our method improves the log likelihood better than APF despite using less than 50\% of the samples. It either performs as good as or slightly better than APF with regards to the deviation of improvement. 

\begin{figure*}
	\centering
	\subfloat{\includegraphics[width=3.25cm, height=3.25cm, clip=true, trim=1.5cm 6.5cm 1.6cm 6.6cm]{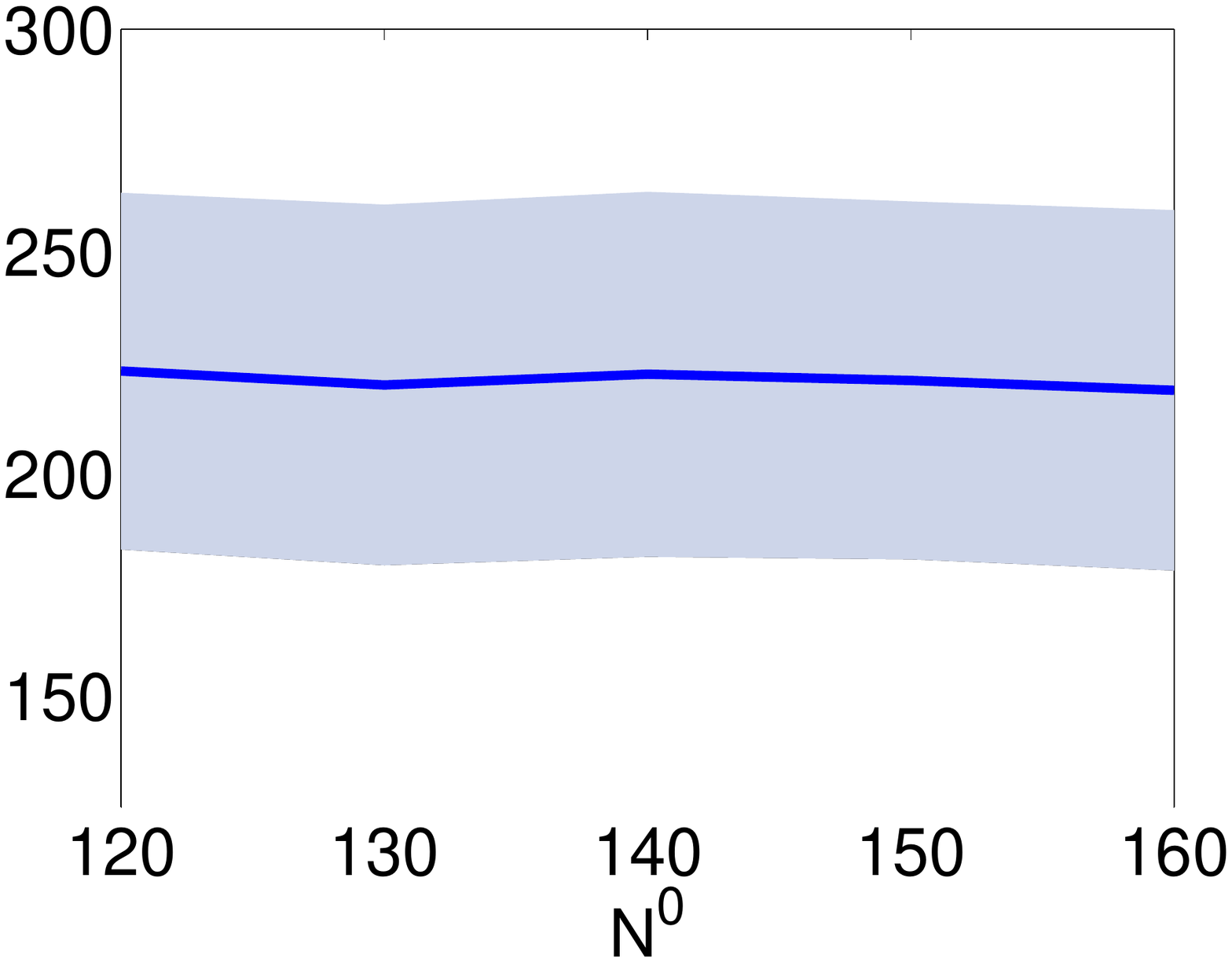}}
	\subfloat{\includegraphics[width=3.25cm, height=3.25cm, clip=true, trim=1.5cm 6.5cm 1.6cm 6.6cm]{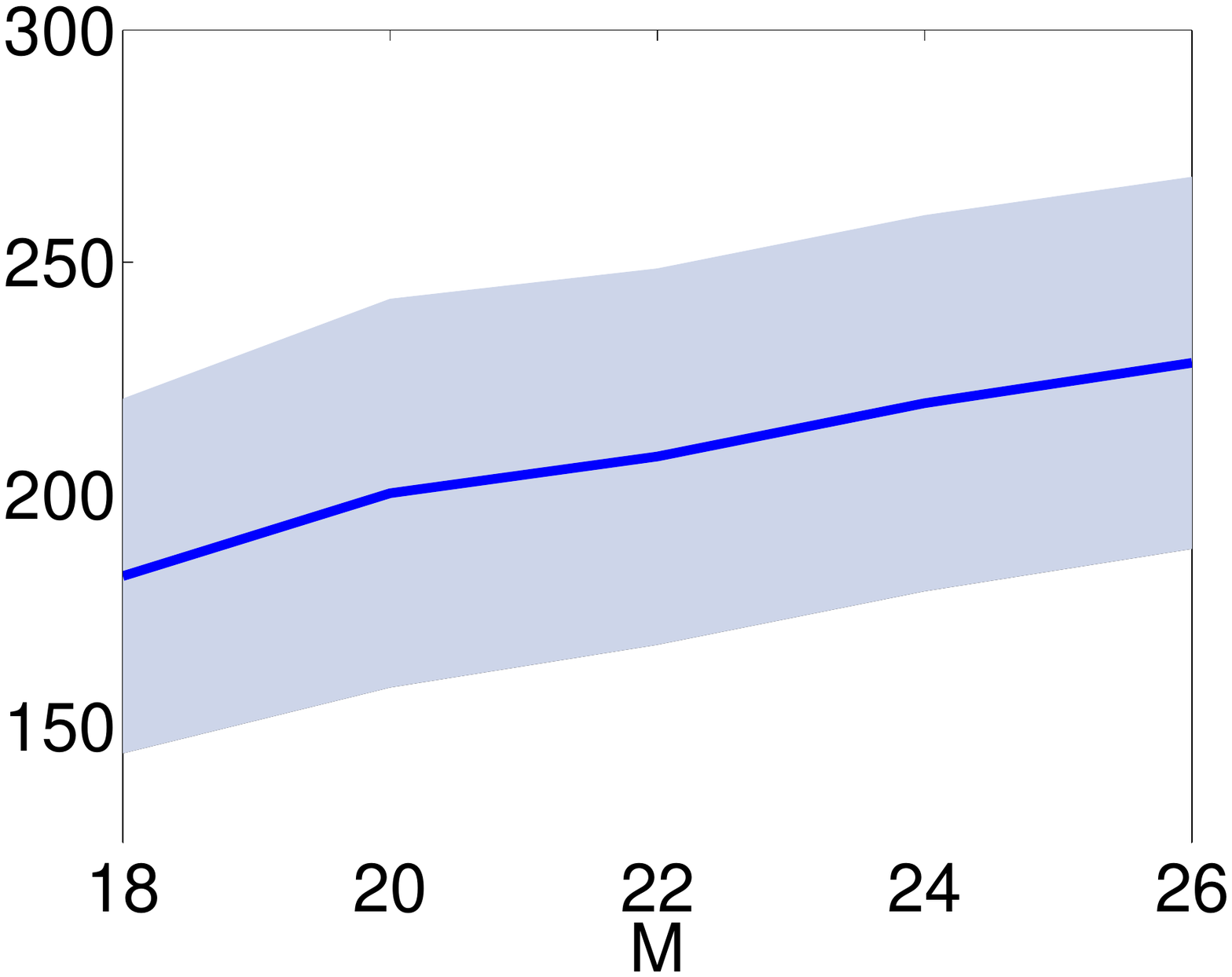}}
	\subfloat{\includegraphics[width=3.25cm, height=3.25cm, clip=true, trim=1.5cm 6.5cm 1.6cm 6.6cm]{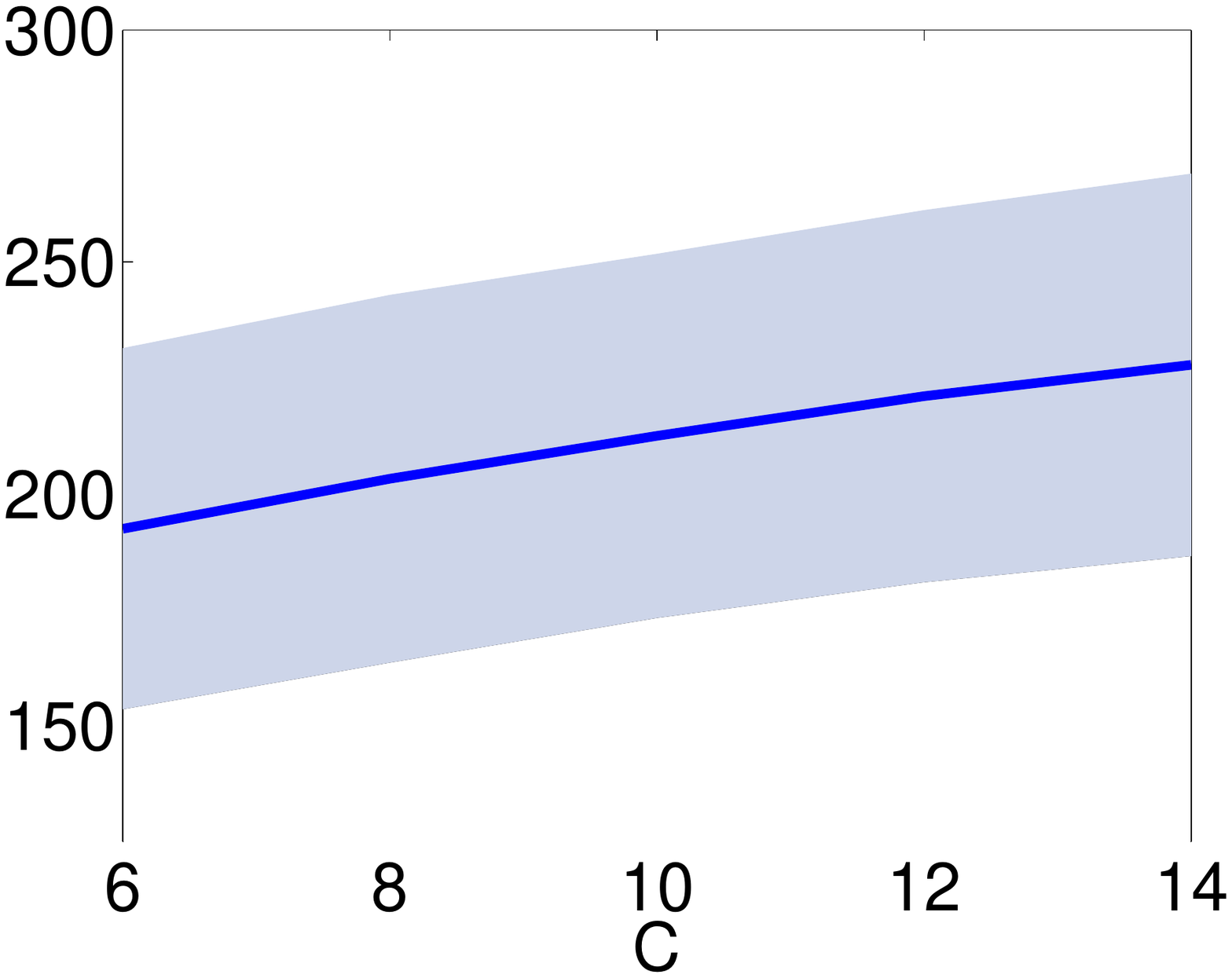}}
	\subfloat{\includegraphics[width=3.25cm, height=3.25cm, clip=true, trim=1.5cm 6.5cm 1.6cm 6.6cm]{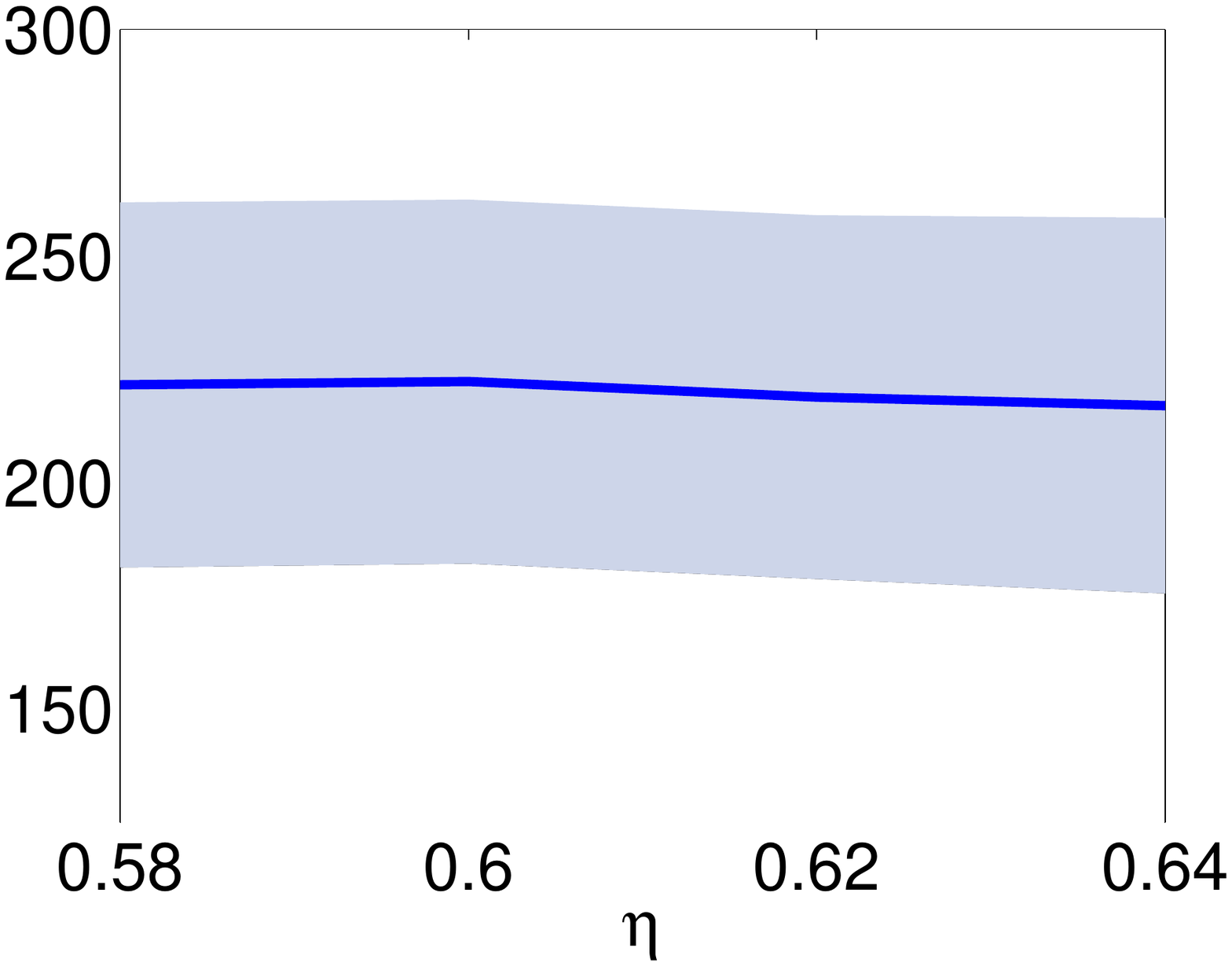}}
	\subfloat{\includegraphics[width=3.25cm, height=3.25cm, clip=true, trim=1.5cm 6.5cm 1.6cm 6.6cm]{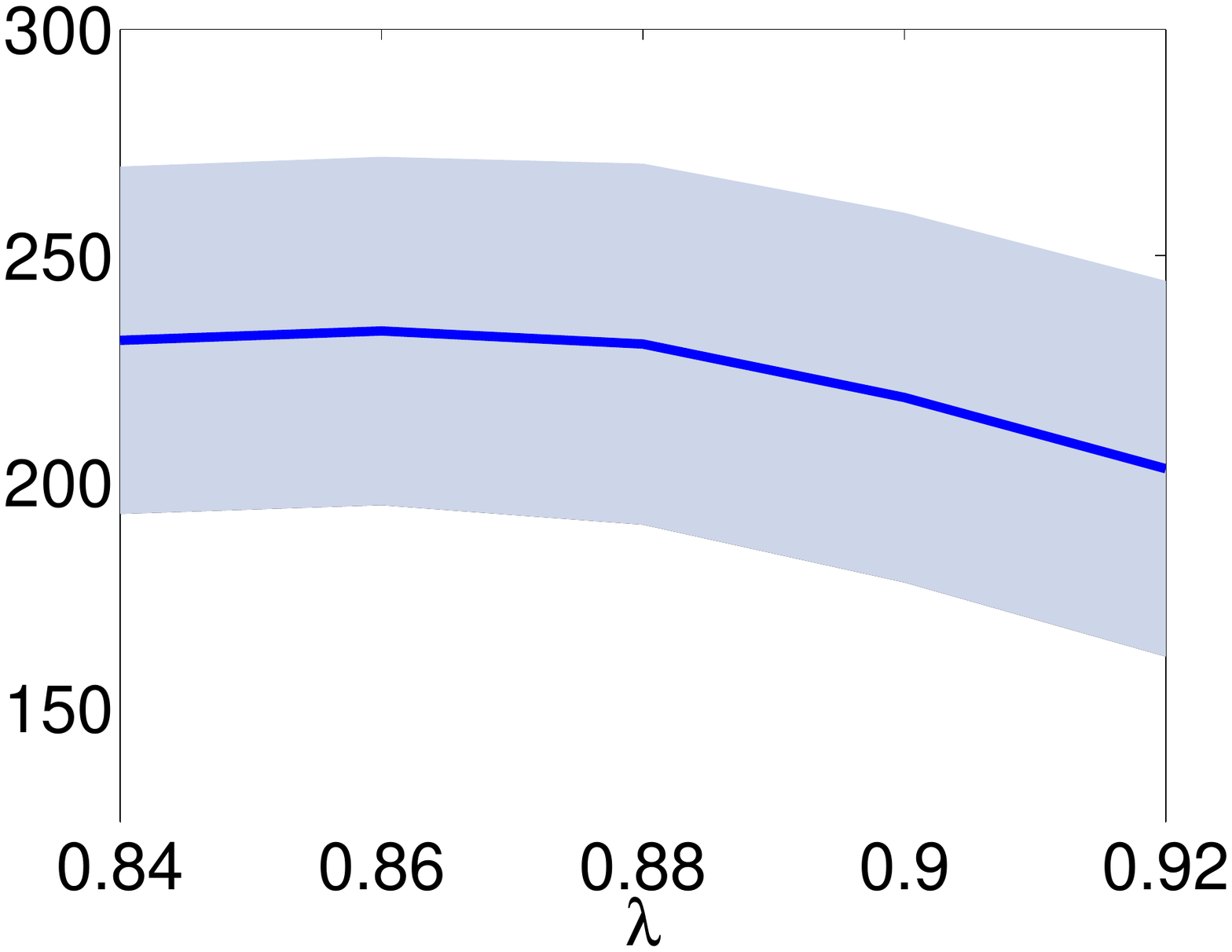}}
	\\
	\caption[Sensitivity analysis for pa]{The plot shows the mean improvement and deviation for various settings of the parameters of our procedure. The parameters $N^0$, $M$ and $C$ refer to initial number of samples, number of annealing layers and the number of new samples introduced per layer respectively. The parameters $\eta$ and $\lambda$ define the power law used to control the rate of annealing. }
	\label{fig:sensanal}
\end{figure*}

\subsection{Sensitivity analysis}
Both our method, as well as APF, have quite a number of parameters. A standard set of parameters proposed in \cite{ApfIjcv} is used in most studies involving APF \cite{HumanEvaII}. However, how these parameters affect the tracking performance has not been studied. This motivated us to perform sensitivity analysis on the parameters of our own algorithm since it would provide insight into how those parameters affect tracking. We performed this study in the same objective as the previous experiment. The test was performed by varying each parameter $\eta$, $\lambda$, $N^0$, $M$ and $C$ around a small range from the default parameter used in previous test and observing the mean and deviation of the variable $I$. Figure \ref{fig:sensanal} shows the results of sensitivity analysis. It can be observed that our algorithm is largely insensitive to the initial number of samples $N^0$. In fact, it slightly improves as the initial number of samples reduce, which may be understood by analyzing Figure \ref{fig:annealexp}. When the initial number of samples are higher, the search is ineffective since the samples hold back the search procedure until they are annealed to a suitable temperature. Similar observations were made with human pose tracking, suggesting one might reduce the initial number of samples without losing performance. 

An increase in the number of annealing layers $M$ improves the performance, this justifies our choice of slow annealing. However it can be observed the deviation shows a slight decreasing trend as the number of layers reduce. We believe this is due to the cumulative effect of the randomness introduced in every layer. Hence a suitable trade-off between the two is necessary. An increase in the number of new samples $C$ introduced in a layer improves the performance, this is expected since when there are more samples the search should be better. A decrease in the parameters $\eta$ and $\lambda$ that define the annealing schedule shows an improvement in performance, indicating that the deeper the annealing, the better the performance of the search procedure.  This is expected since, as shown in Section \ref{sec:motivation}, a deep annealing schedule is necessary to ensure that the new samples search the state space effectively.

\subsection{Tracking Results}
We compared the proposed method with two configurations of APF. 
\begin{inparaenum}[\itshape a\upshape)]
\item 1000 particle configuration with 200 particles per layer and 5 layers
\item 500 particle configuration with 100 particles per layer and 5 layers.
\end{inparaenum}
Our method was configured with 438 samples ($N^0=150, M=24, C=12$). We used data from the Human Eva I dataset for the comparison. We review the overall results here which will be presented in detail elsewhere. Table \ref{tab:evalactobj} presents the results of the comparison from 6 different video sequences. Each tracking method was executed 10 times on a sequence and the time and ensemble average of the tracking error and the deviation are shown in the Table. We observe that, for all videos except one, our method has the lowest tracking error. Furthermore, the difference is significant for jogging sequences.  It is notable that our method reduce despite the fact that it uses half the number of sample, and requires half the runtime as APF 1000.

\begin{figure*}
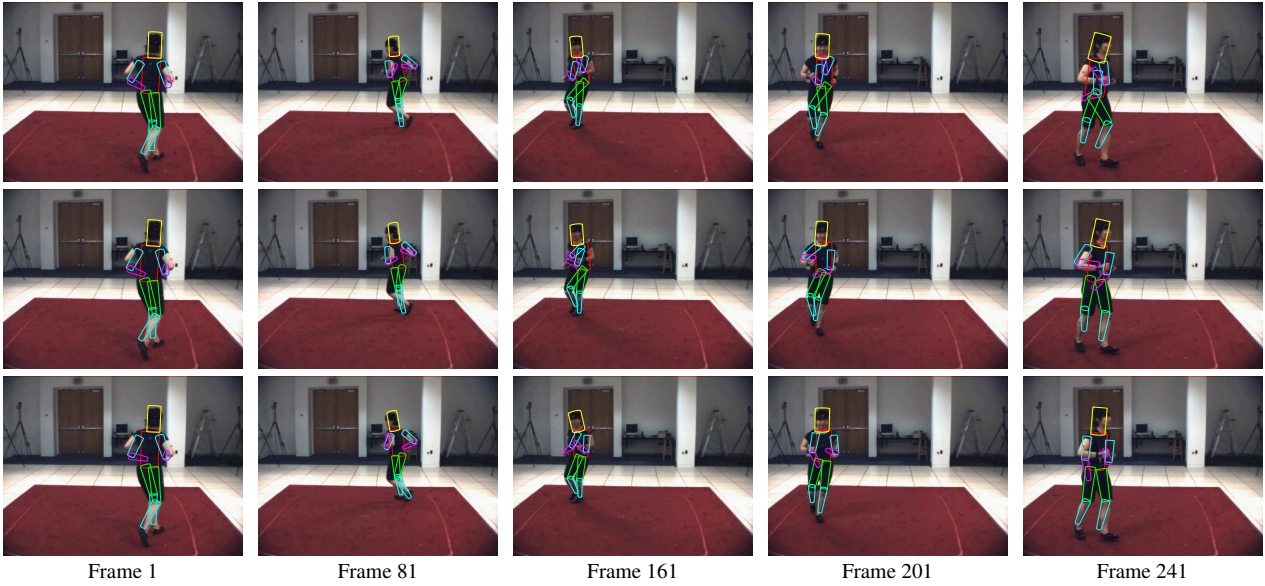

	\captionsetup[subfigure]{labelformat=empty}
	\centering
	\input trackedseqfull.tex
	\caption[Tracked output]{Qualitative comparison of tracked output on Subject 1 jogging sequence. First, second and third row correspond to APF 500, APF 1000 and PA 438 respectively. The model fit can be observed to be worst for APF 500, better for APF 1000, and best for PA 438.}
	\label{fig:trackout}
\end{figure*}

Figure \ref{fig:trackout} shows the tracked model from all three methods superimposed on images used for tracking. We used the Subject 1 jogging sequence for the figure since, as evident from Table \ref{tab:evalactobj} it is the sequence that by a large extent discriminates the three configurations. It can be observed that the tracked output from our method is visibly closer to the Subject.

\begin{table*}
	\begin{center}
    \begin{tabular}{|c|c|c|c|c|c|c|c|c|}
    \hline
    & \thdr{Number} & \multicolumn{6}{|c|}{\thdr{Time averaged mean $\pm$ standard deviation of the tracking error (mm)}} & \thdr{Runtime} \\ 
    \cline{3-8}
    \thdr{Method} & \thdr{of} & \multicolumn{2}{|c|}{\thdr{Subject 1}} & \multicolumn{2}{|c|}{\thdr{Subject 2}} & \multicolumn{2}{|c|}{\thdr{Subject 3}} & \thdr{per} \\
    \cline{3-8}
    & \thdr{Samples} & \thdr{Walk} & \thdr{Jog} & \thdr{Walk} & \thdr{Jog} & \thdr{Walk} & \thdr{Jog} & \thdr{frame} \\
    \hline
				\tcnt{APF} & \tcnt{1000} & {\tcnt{119.2}} $\pm$ {\tcnt{25.6}} & {\tcnt{179.8}} $\pm$ {\tcnt{18.8}} & {\thdr{106.3}} $\pm$ {\tcnt{16.7}} & {\tcnt{122.6}} $\pm$ {\thdr{7.4}} & {\tcnt{132.7}} $\pm$ {\thdr{3.9}} & {\tcnt{115.4}} $\pm$ {\thdr{10.2}} &\tcnt{32.49 s} \\
		\hline
		\tcnt{APF} & \tcnt{500} & {\tcnt{132.4}} $\pm$ {\tcnt{31.5}} & {\tcnt{200.7}} $\pm$ {\tcnt{29.7}} & {\tcnt{107.3}} $\pm$ {\tcnt{16.4}} & {\tcnt{141.6}} $\pm$ {\tcnt{29.2}} & {\tcnt{134.0}} $\pm$ {\tcnt{5.2}} & {\tcnt{127.2}} $\pm$ {\tcnt{21.1}} &\tcnt{16.30 s} \\
		\hline
		\tcnt{PA} & \thdr{438} & {\thdr{93.8}} $\pm$ {\thdr{12.4}} & {\thdr{142.4}} $\pm$ {\thdr{14.5}} & {\tcnt{110.2}} $\pm$ {\thdr{15.3}} & {\thdr{114.4}} $\pm$ {\tcnt{8.9}} & {\thdr{132.3}} $\pm$ {\tcnt{4.4}} & {\thdr{109.8}} $\pm$ {\tcnt{12.2}} &\thdr{14.18 s} \\
		\hline

    \end{tabular}
	\end{center}
	\setlength{\abovecaptionskip}{.1mm}
	\caption{Overall statistics. Best results are shown in bold.}
	\label{tab:evalactobj}
\end{table*}

\section{Conclusion \& Future Work}
In this paper we described a procedure that improves on APF for tracking a human from a video sequence. Using synthetic examples we demonstrate the critical problems with APF and show how they are overcome by the novel aspects of our  procedure, which include reusing samples with a suitably deep annealing schedule, and also by inferring a parametric form from the samples. We then compared the proposed method to APF in a simple scalable problem and show that our method consistently performs better than APF. This was followed by sensitivity analysis showing that our algorithms' performance is largely insensitive to the parameter settings.  Finally, we present human pose tracking results using data from the Human Eva I dataset that show the benefit of using our algorithm.  We plan to explore techniques to optimally estimate the parameters for tracking and to understand how various attributes like the frame rate and the articulated motion affect the tracking performance. 

{\small
\bibliographystyle{ieee}
\bibliography{bibliography}
}

\end{document}